%% file: main.tex
\pdfoutput=1

\documentclass[11pt]{article}

\usepackage{acl}

\usepackage{times}
\usepackage{latexsym}

\usepackage[T1]{fontenc}

\usepackage[utf8]{inputenc}

\usepackage{microtype}

\usepackage[mode=buildnew]{standalone}
\usepackage{algorithm}
\usepackage{algorithmic}
\usepackage{booktabs}
\usepackage{subcaption}
\usepackage{xspace}
\usepackage{enumitem}
\usepackage{pifont}
\usepackage{tikz}
\usepackage{arydshln}

\usepackage{natbib}
\usepackage{amsmath}
\usepackage{hyperref}
\usepackage{amsmath}
\usepackage{amssymb}
\usepackage{mathtools}
\usepackage{multirow}
\usepackage{url}
\usepackage{subcaption}
\usepackage{todonotes}
\usepackage{indentfirst}

\input{utils/math_commands.tex}

\input{utils/macro.tex}

\title{Learning from Executions for Semantic Parsing}

\author{Bailin Wang, Mirella Lapata \and Ivan Titov \\
  Institute for Language, Cognition and Computation \\ 
  School of Informatics, University of Edinburgh \\
  {\tt bailin.wang@ed.ac.uk, \tt \{mlap, ititov\}@inf.ed.ac.uk}}

\date{}

\newcommand\datapercent{70\%\xspace}
\newcommand\datapercentr{30\%\xspace}

\begin{document}

\maketitle
\begin{abstract}
Semantic parsing aims at translating natural language (NL) utterances onto 
machine-interpretable programs, which can be executed against a real-world environment. The expensive annotation 
of utterance-program pairs has long been acknowledged as a major bottleneck for the deployment of contemporary neural
models to real-life applications. In this work, we focus on the task of semi-supervised
learning where a limited amount of annotated data is available together with many unlabeled NL utterances.
Based on the observation that programs which correspond to NL utterances must be always executable,
we propose to encourage a parser to generate executable programs for \emph{unlabeled} utterances.
Due to the large search space of executable programs, conventional  methods that use approximations  based on beam-search 
such as self-training and top-k marginal likelihood training, do not perform as well. 
Instead, we view the problem of learning from executions from the perspective of posterior regularization
and propose a set of new training objectives. Experimental results  on \overnight and \geo show that our new objectives outperform
conventional methods, 
bridging the gap between semi-supervised and supervised learning.
\end{abstract}

\input{sections/introduction.tex}
\input{sections/related_work.tex}
\input{sections/background.tex}
\input{sections/method.tex}
\input{sections/parser.tex}

\input{sections/experiments.tex}
\input{sections/conclusion.tex}

\paragraph{Acknowledgements}
We would like to thank the anonymous reviewers for their valuable comments. We 
gratefully acknowledge the support of the European Research Council (Titov: ERC StG BroadSem 678254; 
Lapata: ERC CoG TransModal 681760) and the Dutch National Science Foundation (NWO VIDI 639.022.518). 

\bibliography{bib/anthology,bib/rebibed_main,bib/text2sql-datasets,bib/overnight}
\bibliographystyle{acl_natbib}

\appendix
\input{appendix.tex}

\end{document}

%% file: utils/math_commands.tex
\usepackage{amsmath,amsfonts,bm}

\def\eqref#1{equation~\ref{#1}}

\def\1{\bm{1}}

\def\vtheta{{\bm{\theta}}}

\def\vr{{\bm{r}}}

\def\vx{{\bm{x}}}

\def\vz{{\bm{z}}}

\DeclareMathAlphabet{\mathsfit}{\encodingdefault}{\sfdefault}{m}{sl}
\SetMathAlphabet{\mathsfit}{bold}{\encodingdefault}{\sfdefault}{bx}{n}

\newcommand{\E}{\mathbb{E}}

\newcommand{\softmax}{\mathrm{softmax}}

\newcommand{\KL}{D_{\mathrm{KL}}}

\DeclareMathOperator*{\argmin}{arg\,min}

%% file: utils/macro.tex
\newcommand{\overnight}{\textsc{Overnight}\xspace}
\newcommand{\geo}{\textsc{GeoQuery}\xspace}
\newcommand{\geoshort}{\textsc{Geo}\xspace}
\newcommand{\method}{\textsc{X-PR}\xspace}

\newcommand{\sfirst}{\textsc{se}}
\newcommand{\ssecond}{\textsc{sn}}
\newcommand{\sthird}{\textsc{ue}}
\newcommand{\sfourth}{\textsc{un}}

\newcommand{\loss}{\mathcal{L}}

\newcommand{\st}{\text{ST}}
\newcommand{\topk}{\text{top-k}}
\newcommand{\lsecond}{\text{repulsion}}
\newcommand{\lthird}{\text{gentle}}
\newcommand{\lfourth}{\text{sparse}}

\newcommand{\lst}{\text{Self-Traing}\xspace}
\newcommand{\ltopk}{\text{Top-K MML}\xspace}
\newcommand{\llsecond}{\text{Repulsion MML}\xspace}
\newcommand{\llthird}{\text{Gentle MML}\xspace}
\newcommand{\llfourth}{\text{Sparse MML}\xspace}

\newcommand{\cmark}{\ding{51}}%
\newcommand{\xmark}{\ding{55}}%

\newcommand{\obj}{\mathcal{J}}
\newcommand{\validp}{\mathcal{Q}}
\newcommand{\validlf}{\mathcal{V}}
\newcommand{\entropy}{\mathcal{H}}
\newcommand{\sparsemax}{\operatorname{SparseMax}}

\newcommand{\comment}[1]{}

%% file: sections/introduction.tex
\section{Introduction}
\label{sec:introduction}

Semantic parsing is the task of mapping natural language (NL) utterances
to meaning representations (aka programs) that can be executed against a real-world
environment such as a knowledge base or a relational database. While neural 
sequence-to-sequence models~\cite{dong-lapata-2016-language,jia-liang-2016-data}
have achieved much success in this task in recent years, they usually
require a large amount of labeled data (i.e.,~utterance-program pairs) for training.
However, annotating utterances with programs is expensive as it 
requires expert knowledge of meaning representations (e.g.,~lambda calculus, SQLs) and 
the environment against which they are executed (e.g.,~a knowledge base, a relational database).
An alternative to  annotation is to collect  answers (or denotations) of programs, rather 
than programs themselves~\cite{liang-etal-2013-learning,berant-etal-2013-semantic}.
In this work, we focus on the more extreme setting where there are no annotations available for a large number of utterances. This setting resembles a common real-life scenario where massive numbers of
user utterances can be collected when deploying a semantic parser~\cite{iyer-etal-2017-learning}. 
Effectively utilizing the unlabeled data makes it possible for a semantic parser to improve over time without 
human involvement. 

\input{figures/example.tex}

Our key observation is that not all candidate programs for an utterance will be semantically valid.
This implies that only some candidate programs 
can be executed and obtain non-empty execution results.\footnote{
In the rest of this paper, we extend the meaning of `executability', 
and use it to refer to the case where a program is executable and obtains non-empty results.}
As illustrated in Figure~\ref{fig:example}, executability is a weak signal that can
differentiate between semantically valid and invalid programs. 
On unlabeled utterances, we can encourage a parser to
only focus on executable programs ignoring non-executable ones. Moreover, the executability
of a program can be obtained from an executor for free without requiring human effort.
Executability has previously been used to guide the decoding of a semantic parser~\cite{wang2018robust}.
We take a step further to directly use this weak signal for learning from unlabeled utterances. 

To learn from executability, we resort to marginal likelihood training, i.e.,~maximizing
the marginal likelihood of all executable programs for an unlabeled NL utterance.
However, the space of all possible programs is exponentially large, as well as the
space of executable ones. Hence, simply marginalizing over all executable programs is intractable.
Typical approximations use beam search to retrieve a handful of (`seen') programs, which are used 
to approximate the entire space. Using such approximations can lead to  optimization getting trapped 
in undesirable local minima. For example, we observe that encouraging a model to exploit seen 
executable programs hinders exploration and reinforces the preference for shorter programs, as 
discussed in Section~\ref{subsec:analysis}. This happens because shorter programs 
are both more likely to be among `seen' programs (probably due to using locally-normalized autoregressive 
modeling) and more likely to be executable.
To alleviate these issues, we derive three new alternative objectives, relying on a new
interpretation of marginal likelihood training from the perspective of posterior regularization. 
Our proposed objectives encode two kinds of inductive biases: 1) 
\textbf{discouraging seen non-executable programs}, which plays a similar role to encouraging seen executable 
ones but does not share its drawback of hindering exploration;
2) \textbf{encouraging sparsity} among executable programs, which
encourages a parser to only focus on a subset of executable programs by softly injecting a sparsity 
constraint. This is desirable, as there are only one or few correct programs for each utterance (see Figure~\ref{fig:example}), and an accurate parser should assign probability mass only to this subset.
We collectively call  these objectives \method, as a shorthand for Execution-guided Posterior Regularization. 

We conduct experiments on two representative semantic
parsing tasks: text-to-LF (logical form) parsing over a knowledge base and text-to-SQL 
\cite{zelle1996learning} parsing over a relational database. Concretely, we evaluate our methods on the \overnight \cite{wang-etal-2015-building} and \geo
datasets. We simulate the semi-supervised learning setting by treating \datapercent of the training data
as unlabeled. Empirical results show that our method can substantially boost the performance of a parser,
trained only on labeled data, by utilizing a large amount of unlabeled data. 

Our contributions are summarized as follows:
%
\begin{itemize}[label={$\bullet$}, topsep=1pt, itemsep=1pt]
    \item We show how to exploit unlabeled utterances by taking advantage of their executability.
    \item 
        To better learn from executability,
        we propose a set of new objectives based on posterior regularization.
    \item 
        Our method can help a base parser achieve substantially better performance 
        by utilizing unlabeled data.
\end{itemize}
Our code, datasets, and splits are publicly available at \url{https://github.com/berlino/tensor2struct-public}.

%% file: figures/example.tex
\begin{figure}
    \setlength{\tabcolsep}{2pt} 
    \begin{tabular}{lcc}
        \toprule 
        list all 3 star rated thai restaurants &	\\
        \midrule
        \textbf{Program Candidates}  & \small{Gold} & \small{Exe} \\
        \textit{select} restaurant \textit{where} star\_rating = thai & {\color{red} \xmark} & {\color{red} \xmark}\\
        \textit{select} restaurant \textit{where} cuisine $>$ 3 & {\color{red} \xmark} & {\color{red} \xmark}\\
        \textit{select} restaurant \textit{where} star\_rating = 3 & {\color{red} \xmark} & {\color{blue} \cmark} \\
        \textit{select} restaurant \textit{where} star\_rating = 3 \\ \quad \textit{and} cuisine = thai &  {\color{blue} \cmark} & {\color{blue} \cmark}  \\
        \bottomrule
    \end{tabular}
    \caption{Candidate programs for an utterance can be classified by executability (Exe); note that 
    the gold program is always in the set of executable programs. We propose to ultilize the weak yet freely available  
    signal of executablility for learning.}
    \label{fig:example}
\end{figure}

%% file: sections/related_work.tex
\section{Related Work}
\label{sec:related_work}

\paragraph{Semi-Supervised Semantic Parsing}
In the context of semantic parsing, semi-supervised models using limited amounts 
of parallel data and large amounts of unlabeled data
treat either utterances or programs as discrete latent variables and induce them
in the framework of generative models~\cite{kocisky-etal-2016-semantic,yin-etal-2018-structvae}.
A challenge with these methods is that (combinatorially) complex
discrete variables make optimization very hard, even with the help of variational inference.
In this work, we seek to directly constrain the discriminative parser with signals obtained from 
executions. Our method can potentially be integrated into these generative models to 
regularize discrete variables.

\paragraph{(Underspecified) Sequence-Level Rewards}
\label{para:seq_reward}
There have been  attempts in recent years to integrate sequence-level rewards  into 
sequence-to-sequence training as a way of accommodating task-specific objectives.  
For example, BLEU can be optimized for coherent text generation~\cite{bosselut2018discourse} 
and machine translation~\cite{wu-etal-2018-study} via reinforcement learning or beam-search~\cite{wiseman2016sequence}. 
In this work, we resort to marginal likelihood training to exploit binary executability rewards for semantic parsing (i.e., whether a program is executable or not), 
which has been shown to be more effective than REINFORCE~\cite{guu-etal-2017-language}.

More importantly, our binary reward is underspecified, i.e., 
there exist many spurious programs that enjoy the same reward as the gold program.
This issue of learning from underspecified rewards underlies many weakly-supervised tasks, 
e.g., learning from denotations~\cite{liang-etal-2013-learning,berant-etal-2013-semantic},
weakly supervised question answering~\cite{min-etal-2019-discrete}.  
Previous work tried to model latent alignments~\cite{wang-etal-2019-learning-semantic} 
between NL and programs to alleviate this issue. In this work, we take an orthogonal direction 
and propose several training objectives that alleviate the impact of spurious programs.

\paragraph{Execution for Semantic Parsing}

Execution has been utilized in semantic parsing~\cite{wang2018robust} and 
the related area of program synthesis~\cite{chen2018execution}. These approaches exploit
the execution of partial programs to guide the search for plausible complete programs. Although 
partial execution is feasible for SQL-style programs, it cannot be trivially extended to general meaning 
representation (e.g., logical forms). In this work, we explore a more general setting where execution can be only 
obtained from complete programs. 

%% file: sections/background.tex
\section{Executability as Learning Signal}

In this section, we formally define our semi-supervised learning setting and show how
to incorporate executability into the training objective
whilst relying on  
the marginal likelihood training framework. We also present two conventional approaches to optimizing  marginal likelihood.


\begin{figure*}[t]
    \centering
    \begin{subfigure}[b]{0.45\textwidth}
        \includestandalone[width=\textwidth]{figures/diagram}
        \caption{Partitioned program space. Red asterisk denotes the most 
            probable executable program $y^*$. P stands for program; subscript S stands for seen, 
            U for unseen, E for executable, and N for non-executable.}
        \label{fig:4parts}
    \end{subfigure}
    \hfill
    \begin{subfigure}[b]{0.45\textwidth}
        \centering
            \begin{align*}
                \loss_{\st} (x, \vtheta) &= - \log p(y^*|x, \vtheta) \\
                \loss_{\topk} (x, \vtheta) &= - \log \sum_{y \in P_{\sfirst}}  p(y|x, \vtheta) \\
                \loss_{\lsecond} (x, \vtheta) &= - \log \big(1 - \sum_{y \in P_{\ssecond}}  p(y|x, \vtheta) \big) \\
                \loss_{\lthird} (x, \vtheta) &= - p(P_\sfirst \cup P_\ssecond) \log \sum_{y \in P_\sfirst}  p(y|x, \vtheta)  \\
                                    & - p(P_\sthird \cup P_\sfourth) \log \sum_{y \in P_\sthird \cup P_\sfourth}  p(y|x, \vtheta) \\
                \loss_{\lfourth} (x, \vtheta) &= - \sum_{y \in P_\sfirst} q_{\lfourth}(y) \log p(y|x, \vtheta)
            \end{align*}
        \caption{Five objectives to approximate MML.}
        \label{fig:objectives_eq}
    \end{subfigure}
    \caption{
        In (a) the program space is partitioned along two dimentions: executability and observability.
        In (b) we show two commonly used objectives (Self-Training and Top-K MML) and the three objectives proposed in this work. 
    }
    \label{fig:objectives}
\end{figure*}

\subsection{Problem Definition}

Given a set of labeled NL-program pairs $\{(x^l_i, y^l_i)\}_{i=1}^N$ 
and a set of unlabeled NL utterances $\{x_j\}_{j=1}^M$, where $N$ and $M$
denote the sizes of the respective datasets, we would like to learn a
neural parser $p(y|x, \vtheta)$, parameterized by $\vtheta$, that maps utterances to programs.
The objective to minimize consists of two parts:
\begin{equation}
    \mathcal J = \frac{1}{N} \sum_{i=1}^{N} \loss_{\text{sup}} (x^l_i, y^l_i) + \lambda \frac{1}{M} \sum_{j=1}^{M} \loss_{\text{unsup}} (x_i)
    \label{eq:obj_semi}
\end{equation}

\noindent where $\loss_{sup}$ and $\loss_{unsup}$ denote the supervised and unsupervised loss, respectively.
For labeled data, we use the  negative log-likelihood of gold programs; for unlabeled data,
we instead use the log marginal likelihood (MML) of \textit{all executable programs}.
Specifically, they are defined as follows:
\begin{align}
    \loss_{\text{sup}}(x,y) &= - \log p(y | x, \vtheta) \\
    \loss_{\text{unsup}}(x) &= - \log \sum_y R(y) p(y | x, \vtheta)
    \label{eq:mml_obj}
\end{align}

\noindent where $R(y)$ is a binary reward function that returns 1 if $y$ 
is executable and 0 otherwise. In practice, this function is 
implemented by running a task-specific executor, e.g., a SQL executor.

Another alternative to unsupervised loss is REINFORCE~\cite{sutton1999policy}, i.e.,
maximize the expected $R(y)$ with respect to $p(y|x, \theta)$. However,
as presented in \citet{guu-etal-2017-language}, this objective usually 
underperforms MML, which is consistent with our initial experiments.\footnote{We 
review the comparison between REINFORCE and MML in the appendix.}

\subsection{Self-Training and Top-K MML}

MML in Equation~(\ref{eq:mml_obj}) requires marginalizing over all executable programs
which is intractable. Conventionally, we resort to beam search to explore the space of programs
and collect executable ones. To illustrate, we can divide the space of programs into four parts 
based on whether they are executable and observed, as shown in Figure~\ref{fig:4parts}. 
For example, programs in $P_\sfirst \cup P_\ssecond$ 
are seen in the sense that they are retrieved by beam search. Programs in $P_\sfirst \cup P_\sthird$ 
are all executable, though only programs in $P_\sfirst$ can be directly observed.

Two common approximations of Equation~(\ref{eq:mml_obj}) are Self-Training (ST) and Top-K MML,
and they are defined as follows:
\begin{align}
    \loss_{\st} (x, \vtheta) &= - \log p(y^*|x, \vtheta) \\
    \loss_{\topk} (x, \vtheta) &= - \log \sum_{y \in P_\sfirst}  p(y|x, \vtheta) 
\end{align}
where $y^*$ denotes the most probable program, and it is approximated by the most probable one
from beam search.

It is obvious that both methods only exploit programs in $P_\sfirst$, i.e., executable programs
retrieved by beam search. In cases where a parser successfully includes the correct programs
in $P_\sfirst$, both approximations should work reasonably well. However, if a parser is uncertain and
 $P_\sfirst$ does not contain the gold program, it would then 
 mistakenly exploit
incorrect programs in learning, which is problematic. 

A naive solution to improve Self-Training or Top-K MML is to explore a larger space, e.g.,~increase 
the beam size to retrieve more executable programs. However, this would inevitably increase the 
computation cost of learning. We also show in the appendix that increasing beam size, after it exceeds a certain 
threshold, is no longer beneficial for learning. In this work, we instead propose better
approximations without increasing beam size.

%% file: figures/diagram.tex
\begin{tikzpicture}
    \draw (0,0) rectangle (4,4);
    \draw (0,2.5) -- (4,2.5);
    \draw (1,4) -- (1,0);

    \node at (0.25, 3.75) {\color{red} *};
    \node at (0.5, 3.25) {$P_{\sfirst}$};
    \node at (0.5, 1.25) {$P_{\ssecond}$};
    \node at (2.5, 3.25) {$P_{\sthird}$};
    \node at (2.5, 1.25) {$P_{\sfourth}$};

    \node[align=center] at (-1.6, 3.25) {\small \emph{Executable} \\ \small \emph{Programs}};
    \node[align=center] at (-1.6, 1.25) {\small \emph{Non-Executable} \\ \small \emph{Programs}};
    \node at (0.2, 4.25) {\small \emph{Seen Programs}};
    \node at (2.85, 4.25) {\small \emph{Unseen Programs}};
\end{tikzpicture}

%% file: sections/method.tex
\section{Method}

We first present a 
view of MML in Equation~(\ref{eq:mml_obj}) from the perspective of posterior regularization.
This new perspective helps us derive three alternative approximations of MML: 
\llsecond, \llthird, and \llfourth.

\subsection{Posterior Regularization}

Posterior regularization (PR) allows to inject linear constraints into 
posterior distributions of generative models, and it can be extended to discriminative 
models~\cite{ganchev2010posterior}. In our case, we try to constrain the parser $p(y|x,\theta)$
to only assign probability mass to executable programs. Instead of imposing hard constraints,
we softly penalize the parser if it is far away from a desired distribution $q(y)$, which is
defined as $\E_q[R(y)] = 1$. Since $R$ is a binary reward function, $q(y)$ is constrained to only place mass on executable
programs whose rewards are 1. We denote all such desired distributions as the family $\validp$.

Specifically, the objective of PR is to penalize the KL-divergence between $\validp$ and $p$, 
which is:
\begin{align}
    \begin{split}
    \obj_{\validp}(\vtheta) &= \KL[ \validp || p(y | x, \vtheta)] \\
        & = \min_{q \in \validp} \KL[ q(y) || p(y | x, \vtheta)]
    \end{split}
    \label{eq:pr_obj}
\end{align}

By definition, the objective has the following upper bound:
\begin{align}
    \begin{split}
    \obj(\vtheta, q) &= \KL[ q(y) || p(y | x, \vtheta)] \\ 
                       &= - \sum_y q(y) \log p(y| x, \vtheta) - \entropy(q)
    \end{split}
    \label{eq:pr_obj_lower}
\end{align}

\noindent where $q \in \validp$, $\entropy$ denotes the entropy.
We can use block-coordinate descent, an EM iterative algorithm to optimize it.
\begin{align*}
    \textsc{E}: q^{t+1} &= \argmin_{q \in \validp} \KL[ q(y) || p(y | x, \vtheta^t)]  \\
    \textsc{M}: \vtheta^{t+1} &= \argmin_{\vtheta} - \sum_{y} q^{t+1}(y)[\log p(y|x,\vtheta)]
\end{align*}

During the E-step, we try to find a distribution $q$ from the constrained set $\validp$ that is closest
to the current parser $p$ in terms of KL-divergence. We then use $q$ as 
a `soft label' and minimize the cross-entropy between $q$ and $p$ during the M-step. 
Note that $q$ is a constant vector and has no gradient wrt. $\vtheta$ during the M-step.

The E-step has a closed-form solution: 
\begin{align}
    \begin{split}
    q^{t+1}(y)  = \left\{ \begin{array}{cc} 
            \frac{p(y|x, \vtheta^t)}{p(P_\sfirst \cup P_\sthird)} & \hspace{1mm} y \in P_\sfirst \cup P_\sthird \\
            0 & \hspace{1mm} \textrm{otherwise} \\
            \end{array} 
    \right.
    \end{split}
    \label{eq:e_step_solution}
\end{align}
where $p(P_\sfirst \cup P_\sthird) = \sum_{y' \in P_\sfirst \cup P_\sthird} p(y'|x, \vtheta^t)$.
$q^{t+1}(y)$ is essentially a re-normalized version of $p$ over executable programs. 
Interestingly, if we use the solution in the M-step, the gradient wrt. $\vtheta$
is equivalent to the gradient of MML in Equation~(\ref{eq:mml_obj}). That is,
optimizing PR with the EM algorithm is equivalent to optimizing MML.\footnote{Please 
see the appendix for the proof and analysis of PR.} The connection between EM and MML is not new, and 
it has been well-studied for classification problems~\cite{amini2002semi,grandvalet2005semi}. 
In our problem, we additionally introduce PR to accommodate the executability constraint, and
instantiate the general EM algorithm.

Although the E-step has a closed-form solution, computing $q$ is still intractable due to 
the large search space of executable programs. However, this PR view provides new insight
on what it means to approximate MML. In essence,
conventional methods can be viewed as computing an approximate solution of $q$. 
Specifically, Self-Training corresponds to a delta distribution that only focuses 
on the most probable~$y^*$.
\begin{align*}
    \begin{split}
    q^{t+1}_{\st}(y) & = \left\{ \begin{array}{cc} 
            1 & \hspace{1mm} y = y^* \\
            0 & \hspace{1mm} \textrm{otherwise} \\
            \end{array} 
    \right.
    \end{split}
\end{align*}
Top-K MML corresponds to a re-normarlized distribution over $P_{\sfirst}$.
\begin{align*}
    \begin{split}
    q^{t+1}_{\topk}(y) & = \left\{ \begin{array}{cc} 
            \frac{p(y|x, \vtheta^t)}{p(P_\sfirst) } & \hspace{1mm} y \in P_\sfirst \\
            0 & \hspace{1mm} \textrm{otherwise} \\
            \end{array} 
    \right.
    \end{split}
\end{align*}
Most importantly, this perspective leads us to deriving
three new approximations of MML, which we  collectively call  \method.

\subsection{\llsecond and \llthird}

As mentioned previously, Self-Training and Top-K MML should be reasonable approximations in
cases where gold programs are retrieved, i.e.,~they are in the seen executable subset ($P_{\sfirst}$ in Figure~\ref{fig:4parts}).
However, if a parser is uncertain, i.e.,~ beam search cannot retrieve the gold programs, 
exclusively exploiting $P_{\sfirst}$  programs is undesirable. Hence, we consider ways of 
taking unseen executable programs ($P_{\sthird}$ in Figure~\ref{fig:4parts})
into account. Since we never directly observe unseen programs ($P_{\sthird}$ or $P_{\sfourth}$),
our heuristics do not discriminate between executable and non-executable programs ($P_{\sthird} \cup P_{\sfourth}$). 
In other words,  upweighting $P_{\sthird}$ programs will inevitably upweight $P_{\sfourth}$.

Based on the intuition that the correct program is included in either seen executable programs ($P_{\sfirst}$)
or unseen programs ($P_{\sthird}$ and $P_{\sfourth}$), we can simply push a parser away from seen non-executable 
programs ($P_{\ssecond}$).
Hence, we call such method \llsecond. 
Specifically, the first heuristic approximates Equation~(\ref{eq:e_step_solution}) as follows:
\begin{align*}
    \begin{split}
    q^{t+1}_{\lsecond}(y) & = \left\{ \begin{array}{cc} 
            \frac{p(y|x, \vtheta^t)}{1 - p(P_\ssecond) } & \hspace{1mm} y \not \in P_\ssecond \\
            0 & \hspace{1mm} \textrm{otherwise} \\
            \end{array} 
    \right.
    \end{split}
\end{align*}

Another way to view this heuristic is that we distribute the probability mass 
from seen non-executable programs ($P_{\ssecond}$) to other programs. 
In contrast, the second heuristic is more `conservative' about unseen programs as it tends 
to trust seen executable $P_{\ssecond}$ programs more.
Specifically, the second heuristic uses the following approximations to solve the E-step. 
\begin{align*}
    \begin{split}
    q^{t+1}_{\lthird}(y) & = \left\{ \begin{array}{cc} 
            \frac{p(P_{\sfirst \cup \ssecond})}{p(P_{\sfirst})}  p(y|x, \vtheta^t) &\hspace{1mm} y \in P_\sfirst \\
            p(y|x, \vtheta^t) & \hspace{-5mm} y \in P_\sthird \cup P_\sfourth \\
            0 & \hspace{1mm} y \in P_\ssecond  \\
            \end{array} 
    \right.
    \end{split}
\end{align*}

Intuitively, it shifts the probability mass of seen non-executable programs ($P_\ssecond$)
directly to seen executable  programs ($P_\sfirst$). 
Meanwhile, it neither upweights nor downweights  unseen programs.
We call this heuristic \llthird.
Compared with Self-Training and Top-K MML, \llsecond~and \llthird~lead to better exploration
of the program space, as only seen non-executable ($P_\ssecond$) programs are discouraged. 

\subsection{\llfourth}
\label{subsec:sparse_mml}

\llfourth~is based on the intuition that in most cases there is only one or few correct programs among all executable programs. 
As mentioned in Section~\ref{para:seq_reward}, spurious programs that are executable, but do not 
reflect the semantics of an utterance are harmful.  
One empirical evidence from previous work~\cite{min-etal-2019-discrete}
is that Self-Training outperforms Top-K MML for weakly-supervised question answering.
Hence, exploiting all seen executable programs can be sub-optimal.
Following recent work on sparse distributions~\cite{martins2016softmax,niculae2018sparsemap},
we propose to encourage sparsity of the `soft label' $q$. 
Encouraging sparsity is also related to the minimum entropy and low-density separation principles which 
are commonly used in semi-supervised learning~\cite{grandvalet2005semi,chapelle2005semi}.

To achieve this, we first interpret the entropy term $\entropy$ in Equation~(\ref{eq:pr_obj_lower})
as a regularization of $q$. It is known that entropy regularization always results in
a dense $q$, i.e., all executable programs are assigned non-zero probability.
Inspired by SparseMax~\cite{martins2016softmax}, we instead use L2 norm for regularization.
Specifically, we replace our PR objective in Equation~(\ref{eq:pr_obj_lower})
with the following one:
\begin{align*}
    \obj_{\text{sparse}}(\vtheta, q) = - \sum_y q(y) \log p(y| s, \vtheta) + \frac{1}{2} ||q||_2^2  
\end{align*}
where $q \in \validp$. Similarly, it can be optimized by the EM algorithm:
\begin{align*}
    \textsc{E}: q^{t+1} &= 
    \sparsemax_{\validp}(\log p(y|x, \vtheta^t))  \\
    \textsc{M}: \vtheta^{t+1} &= \argmin_{\vtheta} - \sum_{y} q^{t+1}(y)[\log p(y|x,\vtheta)]
\end{align*}
where the top-E-step can be solved by the $\sparsemax$ operator, which denotes 
the Euclidean projection from the vector of 
logits $\log p(y|x, \vtheta^t)$  to the simplex $\validp$.
Again, we solve the E-step approximately.
One of the approximations is to use \mbox{top-k} SparseMax 
which constrain the number of non-zeros of $q$ to be less than $k$.
It can be solved by using a top-k operator and followed by 
 $\sparsemax$~\cite{correia2020efficient}.
In our case, we use beam search to approximate the top-k operator
and the resulting approximation for the E-step
is defined as follows:
\begin{align*}
   q^{t+1}_{\lfourth} &= \sparsemax_{y \in P_\sfirst} \big(\log p(y|x, \vtheta^t) \big)    
\end{align*}
Intuitively, $q^{t+1}_{\lfourth}$ occupies the middle ground between
Self-Training (uses $y^*$ only) and Top-K MML (uses all $P_\sfirst$ programs). 
With the help of sparsity of $q$ introduced by $\sparsemax$, 
the M-step will only promote a subset of $P_\sfirst$ programs.


\paragraph{Summary} We propose three new approximations of MML for learning from executions. 
They are designed to complement Self-Training and Top-~K MML
via discouraging seen non-executable programs and introducing sparsity.
In the following sections, we will empirically show that they are 
superior to Self-Training and Top-K MML for semi-supervised semantic parsing. 
The approximations we proposed may also be beneficial for
learning from denotations~\cite{liang-etal-2013-learning,berant-etal-2013-semantic} and
weakly supervised question answering~\cite{min-etal-2019-discrete}, but we leave this to future work. 

%% file: sections/parser.tex
\input{tables/semisup_results_30p.tex}

\section{Semantic Parsers}

In principle, our \method framework  is model-agnostic, i.e., it can be coupled 
with any semantic parser for semi-supervised learning.
In this work, we use a neural parser that achieves state-of-the-art performance across semantic parsing tasks.
Specifically, we use RAT-SQL~\cite{wang2019rat} which features a relation-aware
encoder and a grammar-based decoder. The parser was originally developed for 
text-to-SQL parsing, and we adapt it to text-to-LF parsing.
In this section, we briefly review the encoder and decoder of this parser.
For more details, please refer to \citet{wang2019rat}.

\subsection{Relation-Aware Encoding}

Relation-aware encoding is originally designed to handle \textit{schema encoding}
and \textit{schema linking} for text-to-SQL parsing. We generalize these two notions 
for both text-to-LF and text-to-SQL parsing as follows:
\begin{itemize}[noitemsep,topsep=0pt]
    \item \textit{enviroment encoding}: encoding enviroments, i.e., a knowledge base consisting of a set of triples; a relational database represented by its schema
    \item \textit{enviroment linking}: linking mentions to intended elements of environments, i.e., mentions of entities and properties of knowledge bases; mentions of tables and columns of relational databases
\end{itemize}
%
Relation-aware attention is introduced to inject discrete relations between environment items, 
and between the utterance and environments into the self-attention mechanism of Transformer~\cite{devlin2018bert}. 
The details of relation-aware encoding can be found in the appendix.

\subsection{Grammar-Based Decoding}

Typical sequence-to-sequence models~\cite{dong-lapata-2016-language,jia-liang-2016-data}
treat programs as sequences, ignoring their internal structure. As a result,
the well-formedness of generated programs cannot be guaranteed.
Grammar-based decoders aim to remedy this issue.
For text-to-LF parsing, we use the type-constrained decoder proposed by \citet{krishnamurthy-etal-2017-neural};
for text-to-SQL parsing, we use an AST (abstract syntax tree) based decoder following \citet{yin-neubig-2018-tranx}.
Note that grammar-based decoding can only ensure the syntactic correctness of generated programs.
Executable programs are additionally semantically correct.
For example, all programs in Figure 1 are well-formed, but the first two programs 
are semantically incorrect.

%% file: tables/semisup_results_30p.tex
\begin{table*}[t!]
    \scalebox{0.68}{
        \begin{tabular}{l|cccccccc|c|c}
            \toprule
            & \multicolumn{9}{c|}{\overnight} & \geoshort \\
            Model & \textsc{Basketball} & \textsc{Blocks} & \textsc{Calendar} 
            & \textsc{Housing} & \textsc{Publications} & \textsc{Recipes} 
            & \textsc{Restaurants} & \textsc{Social} & Avg. \\ 
            \midrule
                Lower bound   &    82.2   &   54.1   &   64.9   &   61.4   &   64.3   &   72.7   &   71.7   &   76.7   &   68.5   &  60.6 \\
            \hline
                        \lst   &   84.7   &   52.6   &   67.9   &   59.8   &   68.6   &   80.1   &   71.1   &   77.4   &   70.3   &  64.2 \\
                      \ltopk   &   83.1   &   55.3   &   68.5   &   56.9   &   67.7   &   73.7   &   69.9   &   76.2   &   68.9   &  61.3 \\
            \hdashline
                   \llsecond   &\bf 84.9  &   56.3   &   70.8   &   60.9   &   70.3   &   79.8   &   72.0   & \bf 78.3 &   71.7   &   64.9 \\  
                    \llthird   &   84.1   &   58.1   &   70.2   & \bf 63.0 &   71.5   &   78.7   &   72.3   &   76.4   &   71.8   &   65.6 \\  
                   \llfourth   &   83.9   &\bf 58.6  & \bf 72.6 &   60.3   & \bf 75.2 & \bf 80.6 & \bf 72.6 &   77.8   &\bf 72.7  &  \bf  67.4 \\
            \hline
                Upper Bound    &   87.7   &   62.9   &   82.1   &   71.4   &   78.9   &   82.4   &   82.8   &   80.8   &   78.6   &   74.2\\
            \bottomrule
        \end{tabular}
    }
    \caption{Execution accuracy of supervised and semi-supervised models on all domains of \overnight and \geo. 
    In semi-supervised learning, 30\% of the original training examples are treated as labeled and the remaining 70\%
    as unlabeled. Lower bound refers to supervised models that only use labeled examples and discard unlabeled ones
    whereas upper bound refers to supervised models that have access to gold programs 
    of unlabeled examples. Avg. refers to the average accuracy of the eight \overnight domains.
    We average runs over three random splits of the original training data for semi-supervised learning.
    }
    \label{tab:result30p}
\end{table*}

%% file: sections/experiments.tex
\section{Experiments}
\label{sec:experiments}

To evaluate \method, we present experiments on semi-supervised semantic parsing.
We also analyze how the objectives  affect the training process.

\subsection{Semi-Supervised Learning}

We simulate the setting of semi-supervised learning on standard text-to-LF and text-to-SQL
parsing benchmarks. Specifically, we randomly sample \datapercentr of the original training 
data as the labeled data, and use the rest \datapercent as the unlabeled data. For text-to-LF parsing,
we use the \overnight dataset~\cite{wang-etal-2015-building}, which has eight different 
domains, each with a different size ranging between 801 and 4,419; for text-to-SQL parsing, 
we use \geo~\cite{zelle1996learning} which contains 880 utterance-SQL pairs. 
The semi-supervised setting is very challenging as leveraging only \datapercentr of 
the original training data would result in only around 300 labeled examples in four domains
of \overnight and also in \geo.

\paragraph{Supervised Lower and Upper Bounds}

As baselines, we train two supervised models. The first one only uses the labeled data (\datapercentr of 
the original training data) and discards the unlabeled data in the semi-supervised setting. 
We view this baseline as a \textit{lower bound} in the sense that any semi-supervised method is 
expected to surpass this. The second one has extra access to gold programs for the unlabeled data 
in the semi-supervised setting, which means it uses the full original training data. We view this 
baseline as an \textit{upper bound} for semi-supervised learning; we cannot expect to approach it as the executability signal is much weaker than direct supervision.
By comparing the performance of the second baseline (upper bound) with previous 
methods~\cite{attn-copy-Jia2016,sp-over-many-kbs-Herzig2017,su-yan-2017-xdomain-paraphrasing}, 
we can verify that our semantic parsers are state-of-the-art. Please refer to the Appendix 
for detailed comparisons. Our main experiments aim to show how the proposed objectives can mitigate the gap between the lower- 
and upper-bound baselines by utilizing \datapercent unlabeled data.

\paragraph{Semi-Supervised Training and Tuning}

We use stochastic gradient descent to optimize Equation~(\ref{eq:obj_semi}). 
At each training step, we sample two batches from the labeled  and unlabeled data, respectively. 
In preliminary experiments, we found that it is crucial to pre-train a parser 
on supervised data alone; this is not surprising as all of the objectives for 
learning from execution rely on beam search which would only introduce noise with an untrained parser.
That is, $\lambda$ in Equation~(\ref{eq:obj_semi}) is set to 0 during initial updates, and is switched to a normal 
value afterwards.

We leave out 100 labeled examples for tuning the hyperparameters.
The hyperparameters of the semantic parser are only tuned for the development of the supervised baselines,
and are fixed for semi-supervised learning. The only hyperparameter we tune in the semi-supervised 
setting is the $\lambda$ in Equation~(\ref{eq:obj_semi}), which controls how much the unsupervised objective 
influences learning. After tuning, we use all the labeled examples for supervised training and use the last checkpoints
for evaluation on the test set. 

\input{figures/analysis.tex}

\subsection{Main Results}

Our experiments evaluate the objectives presented in Figure~\ref{fig:objectives}
under a semi-supervised learning setting. Our results are shown
in Table~\ref{tab:result30p}.

\paragraph{Self-Training and Top-K MML}
First, Top-K MML, which exploits more executable programs than Self-Training, does not 
yield better performance in six domains of \overnight and \geo. This observation is consistent with
\citet{min-etal-2019-discrete} where Top-K MML underperforms Self-Training for weakly-supervised
question answering. 
Self-Training outperforms the lower bound in five domains of \overnight, and on average. In contrast, Top-K MML obtains a similar performance to the lower bound 
in terms of average accuracy.

\paragraph{\method Objectives}
In each domain of \overnight and \geo, the objective that achieves the best performance is always
within \method. In terms of average accuracy in \overnight, all our objectives perform better than 
Self-Training and Top-K MML. Among \method, \llfourth performs best in five domains of \overnight,
leading to a margin of 4.2\% compared with the lower bound in terms of average accuracy.  
In \geo, \llfourth also obtain best performance.

Based on the same intuition of discouraging seen non-executable programs, \llsecond 
achieves a similar average accuracy to \llthird in \overnight. In contrast, \llthird tends to perform better 
in domains whose parser are weak (such as \textsc{housing, blocks}) indicated by their lower bounds.
In \geo, \llthird performs slightly better than \llsecond. Although it does not perform better than \llsecond, 
it retrieves more accurate programs and also generates longer programs (see next section for details).

To see how much labeled data would be needed for a supervised model to 
reach the same accuracy as our semi-supervised models, 
we conduct experiments using 40\% of the original training examples as 
the labeled data. The supervised model achieves 72.6\% on average in \overnight,
implying that ‘labeling’ 33.3\% more  examples would yield the same accuracy 
as our best-performning objective (\llfourth).

\subsection{Analysis}
\label{subsec:analysis}

To better understand the effect of different objectives, we conduct analysis
on the training process of semi-supervised learning. For the sake of brevity, 
we focus our analysis on the \textsc{Calendar} domain but have drawn similar conclusions for the other domains. 

\paragraph{Length Ratio}

During preliminary experiments, we found that all training 
objectives tend to favor short executable programs for 
unlabeled utterances. To quantify this, 
we define the metric of average ratio as follows:
\begin{equation}
    ratio = \frac{\sum_i \sum_{y \in P_\sfirst(x_i)}{|y|}}{\sum_i|x_i||P_\sfirst(x_i)|}
\end{equation}
where $ P_\sfirst(x_i)$ denotes seen executable programs of $x_i$,
$|x|, |y|$  denotes the length of  an utterance and a program, respectively, 
 and $|P_\sfirst(x_i)|$ denotes the number of seen executable programs. Intuitively,
average ratio reveals the range of programs that an objective is 
exploiting in terms of length. This metric is computed in an online 
manner, and $x_i$ is a sequence of data fed to
the training process. 

As shown in Figure~\ref{fig:avg_ratio},  Top-K MML
favors shorter programs, especially during the initial steps. In contrast, 
\llsecond~and \llthird~prefer longer programs.  For reference, we can compute the gold ratio 
by assuming $ P_\sfirst(x_i)$  only contains the gold program. The gold ratio for \textsc{Calendar}
is 2.01, indicating that all objectives are still preferring programs that are shorter than gold programs.
However, by not directly exploiting seen executable programs, \llsecond~and \llthird~ alleviate this issue
compared with Top-K MML.

\paragraph{Coverage}

Next, we analyze how much an objective can help a parser retrieve gold programs for 
unlabeled data. Since the orignal data contains the gold programs for the unlabeled data, 
we ultilize them to define the metric of coverage  as follows:
\begin{equation}
 coverage = \frac{ \sum_i I[\hat y_i \in P_{\sfirst}(x_i)]}{\sum_i|x_i|}
\end{equation}
where $I$ is an indicator function, $\hat y_i$ denotes the gold program
of an utterance $x_i$. Intuitively, this metric measures how often 
a gold program is captured in $P_{\sfirst}$. 
As shown in Figure~\ref{fig:coverage}, Self-Training, which 
only exploits one program at a time, is relatively weak in terms
of retrieving more gold programs. In contrast, \llsecond~
retrieves more gold programs than the others.

As mentioned in Section~\ref{subsec:sparse_mml}, SparseMax can be viewed as an 
interpolation between Self-Training and Top-K MML. This is also reflected in both
metrics: \llfourth~always occupies the middle-ground performance between ST and Top-K MML.   
Interestingly, although \llfourth~is not best in terms of both diagnostic 
metrics, it still achieves the best accuracy in this domain.

%% file: figures/analysis.tex
\begin{figure*}[t]
    \centering
    \begin{subfigure}[b]{0.48\textwidth}
        \includegraphics[width=\linewidth]{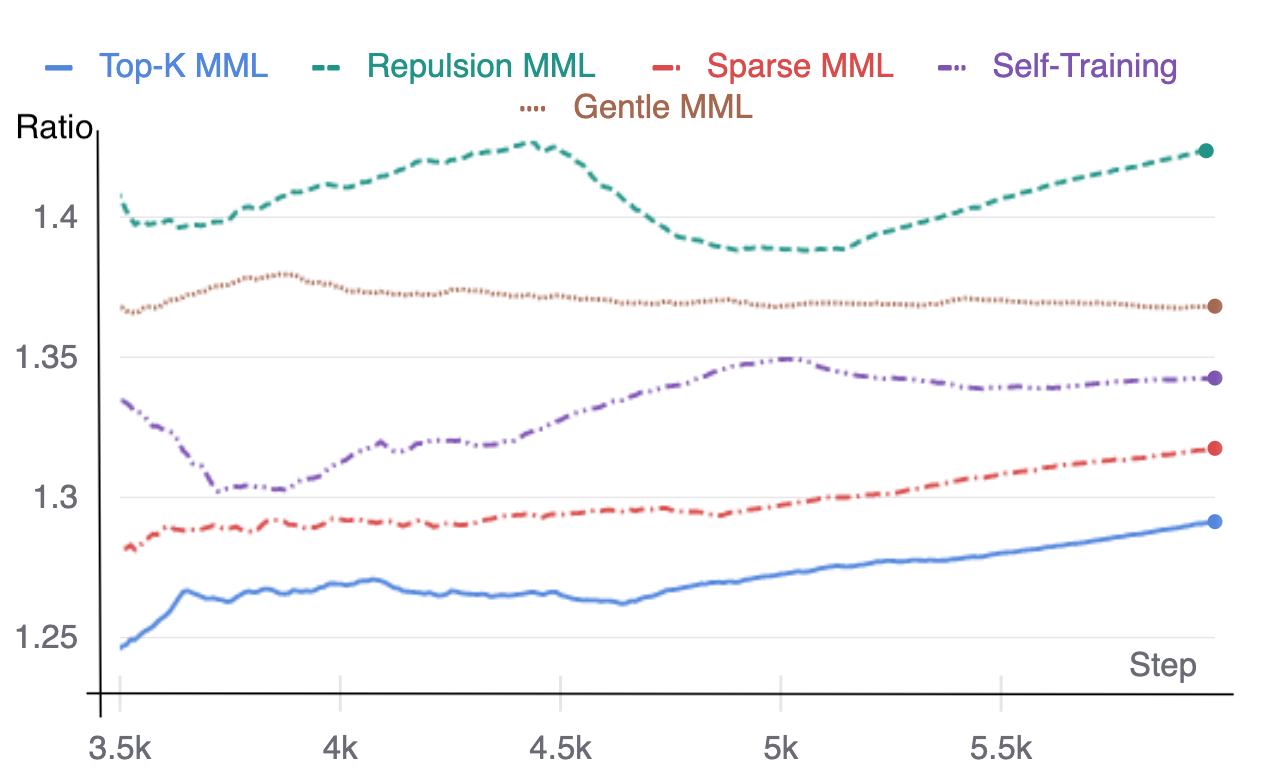}
        \caption{Average Ratios.}
        \label{fig:avg_ratio}
    \end{subfigure}
    \centering
    \begin{subfigure}[b]{0.48\textwidth}
        \includegraphics[width=\linewidth]{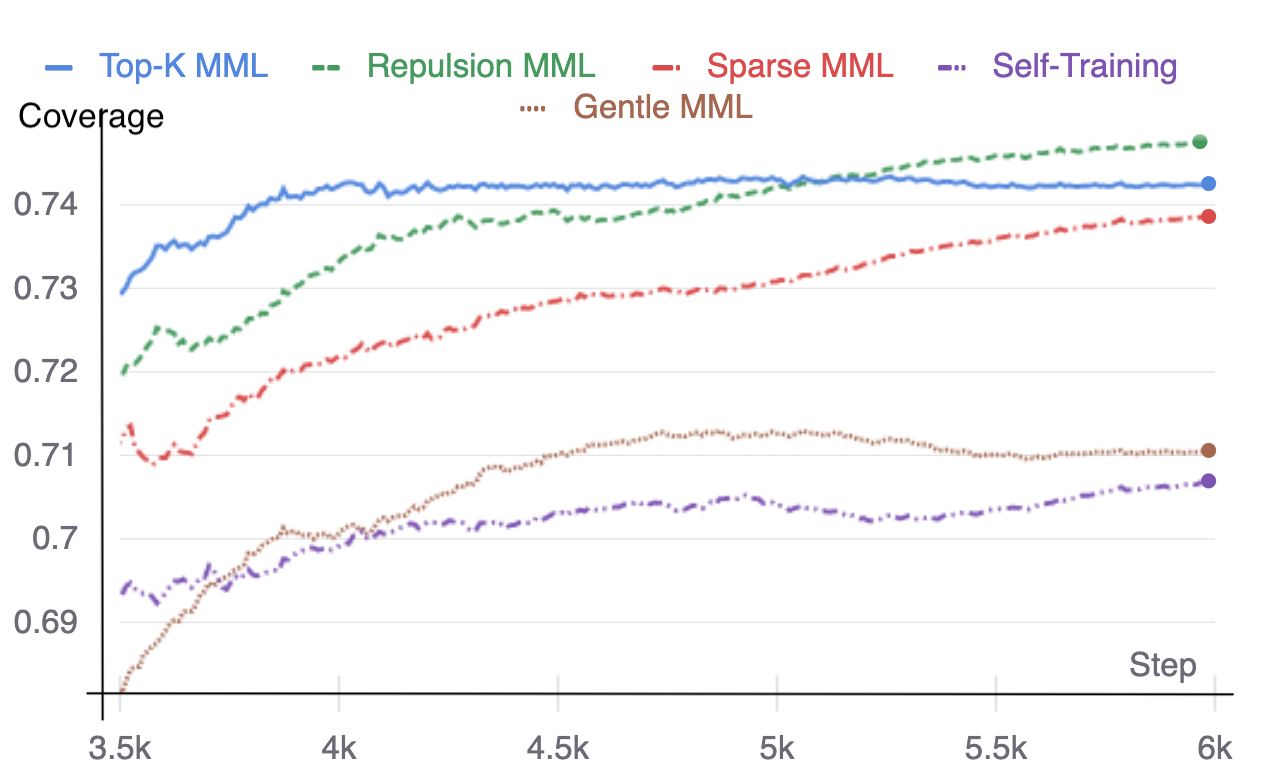}
        \caption{Coverage of gold programs.}
        \label{fig:coverage}
    \end{subfigure}
    \caption{Effect of different learning objectives in terms of average ratios and coverage (view in color).}
\end{figure*}

%% file: sections/conclusion.tex
\section{Conclusion}

In this work, we propose to learn a semi-supervised semantic parser 
from the weak yet freely available executability signals.
Due to the large  search space of executable programs, conventional approximations
of MML training, i.e, Self-Training and Top-K MML, are often sub-optimal. 
We propose a set of alternative objectives, namely \method, through the lens 
of posterior regularization. 
Empirical results on semi-supervised learning show that \method can help a parser achieve substantially better performance 
than conventional methods, further bridging the gap between semi-supervised learning and supervised learning. 
In the future, we would like to extend \method to related tasks such as learning from denotations 
and weakly supervised question answering.

%% file: appendix.tex
\section{Method}

\paragraph{MML vs. RL}

Another choice for learning from executions is to maximize the expected reward:
\begin{equation}
  \obj_{\text{RL}}(x, \vtheta)  = \E_{p(y|x, \vtheta)} [R(y)]
\end{equation}
Recall that our MML objective is defined as:
\begin{align}
    \obj_{\text{MML}} (x, \vtheta) = \log \sum_y R(y) p(y|x, \vtheta)
    \label{eq:mml}
\end{align}
As shown in previous work~\cite{guu-etal-2017-language}, the gradients (wrt. $\vtheta$)
of RL and MML have the same form:

\begin{align}
    \nabla_\vtheta \obj (x, \vtheta) = \sum_y q(y) \nabla_\vtheta \log p(y|x, \vtheta)
    \label{eq:mml_grad}
\end{align}
where $q$ can be viewed as a soft-label that is induced from $p$.
For RL, $q_{\text{RL}}(y) = p_\vtheta(y|x, \theta)R(y)$; 
for MML, $q_{\text{MML}}(y)=\frac{R(y)p_\vtheta(y|x)}{\sum_{y'}R(y')p_\vtheta(y'|x)} = p_\vtheta(y|x, R(y)=1)$.
Compared with RL, MML additionally renormalize the $p$ over executable programs to obtain $q$.
As a result, MML outputs a `stronger' gradient than RL due to the renormalization.
In our initial experiments, we found that RL is hard to converge whereas MML is not.

\paragraph{Proof of the E-Step Solution}
Denote the set of executable programs as $\validlf$. 
Since $q(y)$ is 0 for non-executable programs, we only need to compute it for executable programs in $\validlf$. 
We need to solve the following optimization problem:

\begin{align}
    \begin{split}
        \min_q & - \sum_{y \in \validlf} q(y) \log p(y|x, \vtheta) + \sum_{y \in \validlf} q(y) \log q(y) \\
            s.t. \quad & \sum_{y \in \validlf} q(y) = 1 \\
            & q(y) \geq 0 
    \end{split}
    \label{eq:kkt_obj}
\end{align}

By solving it with KKT conditions, we can see that $q(y) \propto p(y|x,\vtheta)$.
Since $q(y)$ needs to sum up to 1, it is easy to obtain that $q(y) = \frac{p(y|x, \vtheta)}{\sum_{y \in \validlf}p(y|x, \vtheta)}$.


\section{Experiments}

\input{tables/rat_relations.tex}
\paragraph{Relation-Aware Encoding of Semantic Parsers}
Relation-aware encoding was originally introduced for text-to-SQL parsing 
in~\citet{wang2019rat} to accommodate discrete relations among schema items (e.g., tables and columns)
and linking between an utterance and schema items.
Let $\{\vx_i^{(0)}\}_{i=0}^{N-1}$ denote the input to the parser consisting of NL tokens and the
(linearized version of) environment; relation-aware encoding changes the multi-head self-attention
(with $H$ heads and hidden size $d_x$) as follows:
\begin{equation}
    \label{eq:rat_encoding}
    \begin{aligned}
        e_{ij}^{(n,h)} &= \frac{\vx^{(n)}_i W_Q^{(n,h)} (\vx^{(n)}_j W_K^{(n,h)} + {\color{red} \vr_{ij}^K})^\top}{\sqrt{d_z/ H}}
        \\
        \alpha_{ij}^{(n,h)} &= \softmax_{j} \bigl\{ e_{ij}^{(n,h)} \bigr\}
        \\
        \vz_i^{(n, h)} &= \sum_{j=1}^n \alpha_{ij}^{(n,h)} (\vx_j^{(n)} W_V^{(n,h)} + {\color{red} \vr_{ij}^V}).
    \end{aligned}
\end{equation}
where $\alpha_{ij}^{(n,h)}$ denotes the attention weights of head $h$ at layer $n$, \mbox{$0 \le h < H$},
\mbox{$0 \le n < N$}, and \\ \mbox{$W_Q^{(h)}, W_K^{(h)}, W_V^{(h)} \in \mathbb{R}^{d_x \times (d_x / H)}$}.
Most importantly, $\vr_{ij}^K$ and $\vr_{ij}^V$ are key and value embeddings of the discrete relation $r_{ij}$ between items $i$ and $j$.
They are incorporated to bias attention towards discrete relations.

In this work, we re-use the relations from \citet{wang2019rat} for our 
text-to-SQL parsing task on the \geo dataset. For text-to-LF parsing on 
the \overnight dataset, we elaborate on how we define discrete relations.
The input of text-to-LF parsing is an utterance and a fixed knowledge base $\mathcal K$ which 
is represented as a set of triples in the form of (entity1, property, entity2).
To feed the input into a transformer, we first linearize it into an ordered sequence which contains
a list of utterance tokens, followed by a sequence of entities and properties. 
To model the relations among the items of this sequence, we define relations
between each pair of items, denoted by ($x$, $y$), in Table~\ref{tab:lf_edges}.

\paragraph{Statitics of Datasets}
\input{tables/data_stat.tex}
The numbers of examples in each domain are shown in Table~\ref{tab:stat}.
Four domains of \overnight and \geo contain only around 1000 examples.

\paragraph{Results of Supervised Models}
\input{tables/sup_results.tex}
The results of supervised models (upper bounds) are shown in 
Table~\ref{tab:sup_results}. Our parser achieves the best performance 
among models without cross-domain training. This confirms that we use 
a strong base parser for our experiments on semi-supervised learning. 

\paragraph{Beam Size}
We try the beam size from $\{4, 8, 16, 32, 64\}$ and finally picks 16 which performs best in
the \overnight dataset. We also try a larger beam size of 128 during preliminary experiments. However, 
the model is extremely slow to train and does not outperform the one with beam size 16.

\paragraph{Semi-Supervised Learning with 10\% Data}

\input{tables/semisup_results_10p.tex}

We also investigate a semi-supervised setting where only 10\% of the original 
training data are treated as labeled data. This is more challenging as 
four domains of \overnight and \geo only have around 100 labeled  utterance-program pairs
in this setting. The results are shown in Table~\ref{tab:result10p}. 
In terms of average accuracy, \llfourth achieves the best performance. 
The margin of improvement, compared to the lower bound, is relatively smaller 
than using 30\% data (3\% vs 4.2\%). As all objectives rely on beam search,
a weak base parser, as indicated by the lower bound, is probably harder to improve.
By taking account of unseen programs, \llsecond and \llthird tend to be more 
useful in domains such as \textsc{Housing, Publications, Recipes} where the 
base parsers are very weak (accuracy is around 40\%). Moreover, \llthird performs best in \geo.

%% file: tables/rat_relations.tex
\begin{table*}[t]
    \centering
    \scalebox{0.8}{
    \begin{tabular}{lllp{7cm}}
        \toprule
        Type of $x$ & Type of $y$ & Edge label & Description \\
        \midrule
        \multirow{2}{*}{Entity} & \multirow{2}{*}{Entity}
   & \textsc{Related-F}    & there exists a property $p$ s.t. $(x, p, y) \in \mathcal K$ \\
 & & \textsc{Related-R} & there exists a property $p$ s.t. $(y, p, x) \in \mathcal K$  \\
 \midrule
 \multirow{2}{*}{Entity} & \multirow{2}{*}{Property}
   & \textsc{Has-Property-F}   & there exists an entity $e$  s.t. $(x, y, e) \in \mathcal K$ \\
 & & \textsc{Has-Property-R}    & there exists an entity $e$  s.t. $(e, y, x) \in \mathcal K$ \\
 \midrule
 \multirow{2}{*}{Property} & \multirow{2}{*}{Entity}
   & \textsc{Prop-to-Ent-F}   & there exists an entity $e$  s.t. $(y, x, e) \in \mathcal K$ \\
 & & \textsc{Prop-to-Ent-R}   & there exists an entity $e$  s.t. $(e, x, y) \in \mathcal K$\\
 \midrule
 \multirow{2}{*}{Utterance Token} & \multirow{2}{*}{Entity}
   & \textsc{Exact-Match}   & $x$ and $y$ are the same word \\
 & & \textsc{Partial-Match}   & token $x$ is contained in entity $y$ \\
 \midrule
 \multirow{2}{*}{Entity} & \multirow{2}{*}{Utterance Token}
   & \textsc{Exact-Match-R}   & $y$ and $x$ are the same word \\
 & & \textsc{Partial-Match-R}   & token $y$ is contained in entity $x$ \\
 \midrule
 \multirow{2}{*}{Utterance Token} & \multirow{2}{*}{Property}
   & \textsc{P-Exact-Match}   & $x$ and $y$ are the same word \\
 & & \textsc{P-Partial-Match}   & token $x$ is contained in property $y$ \\
 \midrule
 \multirow{2}{*}{Property} & \multirow{2}{*}{Utterance Token}
   & \textsc{P-Exact-Match-R}   & $y$ and $x$ are the same word \\
 & & \textsc{P-Partial-Match-R}   & token $y$ is contained in property $y$ \\
 \bottomrule
\end{tabular}
}
\caption{Relation types used for text-to-LF parsing.}
\label{tab:lf_edges}
\end{table*}

%% file: tables/data_stat.tex
\begin{table*}[t]
    \scalebox{0.73}{
        \begin{tabular}{l|cccccccc|c}
            \toprule
             & \textsc{Basketball} & \textsc{Blocks} & \textsc{Calendar} 
            & \textsc{Housing} & \textsc{Publications} & \textsc{Recipes} 
            & \textsc{Restaurants} & \textsc{Social} & \geoshort \\ 
            \midrule
                all   & 1952 & 1995 & 837  & 941  & 801  & 1080 & 1657 & 4419 & 880 \\
            \bottomrule
        \end{tabular}%
    }
    \caption{Data sizes of the \overnight and \geo dataset.}
    \label{tab:stat}
\end{table*}

%% file: tables/sup_results.tex
\begin{table*}[t]
    \scalebox{0.68}{
        \begin{tabular}{l|cccccccc|c}
            \toprule
            Model & \textsc{Basketball} & \textsc{Blocks} & \textsc{Calendar} 
            & \textsc{Housing} & \textsc{Publications} & \textsc{Recipes} 
            & \textsc{Restaurants} & \textsc{Social} & Avg. \\ 
            \midrule
            \citet{Overnight-Wang15} & 46.3 & 41.9 & 74.4 & 54.5 & 59.0 & 70.8 & 75.9 & 48.2 & 58.8 \\
            \citet{attn-copy-Jia2016} & 85.2 & 58.1 & 78.0 & 71.4 & 76.4 & 79.6 & 76.2 & 81.4 & 75.8 \\
            \citet{sp-over-many-kbs-Herzig2017} & 85.2 & 61.2 & 77.4 & 67.7 & 74.5 & 79.2 & 79.5 & 80.2 & 75.6 \\
            \citet{su-yan-2017-xdomain-paraphrasing} & 86.2 & 60.2 & 79.8 & 71.4 & 78.9 & 84.7 & 81.6 & 82.9 & 78.2 \\
            \hdashline
            \bf Ours & 87.7 & 62.9 & 82.1 & 71.4 & 78.9 & 82.4 & 82.8 & 80.8 & \bf 78.6 \\
            \hline
            \citet{sp-over-many-kbs-Herzig2017}$^*$ & 86.2 & 62.7 & 82.1 & 78.3 & 80.7 & 82.9 & 82.2 & 81.7 & 79.6 \\
            \citet{su-yan-2017-xdomain-paraphrasing}$^*$ & 88.2 & 62.7 & 82.7 & 78.8 & 80.7 & 86.1 & 83.7 & 83.1 & 80.8 \\
            \bottomrule
        \end{tabular}%
    }
    \caption{Test accuracy of supervised models on all domains for \overnight. Models with $^*$ are augmented with cross-domain training.}
    \label{tab:sup_results}
\end{table*}

%% file: tables/semisup_results_10p.tex
\begin{table*}[t!]
    \scalebox{0.68}{
        \begin{tabular}{l|cccccccc|c|c}
            \toprule
            & \multicolumn{9}{c|}{\overnight} & \geoshort \\
            Model & \textsc{Basketball} & \textsc{Blocks} & \textsc{Calendar} 
            & \textsc{Housing} & \textsc{Publications} & \textsc{Recipes} 
            & \textsc{Restaurants} & \textsc{Social} & Avg. \\ 
            \midrule
                Lower Bound    &   68.0   &   35.3   &   40.5   &   34.4   &   39.1   &   45.8   &   59.3   &   63.7   &   48.3   &   42.7 \\
            \hline
                        \lst   &   66.5   &   37.8   & \bf 48.2 & \bf 39.2   &   41.0   &   41.2 & \bf  63.0   &   63.8   &   50.1   &   44.1 \\
                      \ltopk   &   70.3   &   36.8   &   42.3   &   30.7   &   37.3   &   44.4   &   61.1   &   62.2   &   48.1   &   40.1 \\
            \hdashline
                   \llsecond   &   70.8   &   37.3   &   44.6   &   37.8  &   39.8   &   44.4   &   60.2   & \bf 67.8 &   50.3   &   44.8 \\  
                    \llthird   &   68.3   &   38.1   &   45.8   &\bf 39.2 & \bf 43.5 & \bf 44.9 &   59.9   &   64.0   &   50.5   & \bf 46.6 \\  
                   \llfourth   & \bf 73.4 & \bf 42.1 &   46.4   &   38.1   &   42.2   &   43.1   & \bf 63.0 &   64.9   &\bf 51.3  &  45.2 \\
            \hline
                Upper Bound    &   87.7   &   62.9   &   82.1   &   71.4   &   78.9   &   82.4   &   82.8   &   80.8   &   78.6   &   74.2\\
            \bottomrule
        \end{tabular}
    }
    \caption{Execution accuracy of supervised and semi-supervised models on all domains of \overnight and \geo. 
    In semi-supervised learning, 10\% of the original training examples are treated as labeled and the remaining 90\% as 
    unlabeled. Lower bound refers to supervised models that only use labeled examples and discard unlabeled ones
    whereas upper bound refers to supervised models that have access to gold programs 
    of unlabeled examples. Avg. refers to the average accuracy of the eight \overnight domains.}
    \label{tab:result10p}
\end{table*}